\documentclass{ecai} 

\usepackage{latexsym}
\usepackage{amssymb}
\usepackage{amsmath}
\usepackage{amsthm}
\usepackage{booktabs}
\usepackage{enumitem}
\usepackage{graphicx}
\usepackage{color}

\usepackage{algorithm}
\usepackage[noend]{algpseudocode}

\usepackage{comment}
\usepackage{amsfonts}       
\usepackage{nicefrac}       

\usepackage{booktabs}

\usepackage{subcaption} 

\usepackage{multibib}

\usepackage{amsmath}
\DeclareMathOperator*{\arm}{arm} 
\DeclareMathOperator*{\armset}{armset} 
\DeclareMathOperator*{\task}{task} 
\DeclareMathOperator*{\bandit}{bandit} 
 
\newcommand{\T}{\mathcal{T}}
\newcommand{\hmu}{\widehat{\mu}}
\newcommand{\hsi}{\widehat{\sigma}}
\newcommand{\hDe}{\widehat\Delta}
\newcommand{\hH}{\widehat{H}}

\newcommand{\BibTeX}{B\kern-.05em{\sc i\kern-.025em b}\kern-.08em\TeX}

\begin{document}


\begin{frontmatter}

\title{BandiK: Efficient Multi-Task Decomposition \\Using a  Multi-Bandit Framework}

\author[A,B]{\fnms{András}~\snm{Millinghoffer}\footnote{Equal contribution.}}
\author[A,C]{\fnms{András}~\snm{Formanek}\footnotemark}
\author[A]{\fnms{András}~\snm{Antos}}
\author[A,B]{\fnms{Péter}~\snm{Antal}\thanks{Corresponding Author. Email: antal@mit.bme.hu}} 

\address[A]{Department of Artificial Intelligence and Systems Engineering, Faculty of Electrical Engineering and Informatics, Budapest University of Technology and Economics, Műegyetem rkp. 3, H-1111 Budapest, Hungary}
\address[B]{ E-Group ICT Software Zrt., Alsó Törökvész út 2, H-1022 Budapest, Hungary    }
\address[C]{Department of Electrical Engineering (ESAT), STADIUS Center for Dynamical Systems, Signal Processing and Data Analytics, KU Leuven, 3001 Leuven, Belgium}

\begin{abstract}

The challenge of effectively transferring knowledge across multiple tasks is of critical importance and is also present in downstream tasks with foundation models. However, the nature of transfer, its transitive-intransitive nature, is still an open problem, and negative transfer remains a significant obstacle. Selection of beneficial auxiliary task sets in multi-task learning is frequently hindered by the high computational cost of their evaluation, the high number of plausible candidate auxiliary sets, and the varying complexity of selection across target tasks. 

To address these constraints, we introduce BandiK, a novel three-stage multi-task auxiliary task subset selection method using multi-bandits, where each arm pull evaluates candidate auxiliary sets by training and testing a multiple output neural network on a single random train-test dataset split. Firstly, BandiK estimates the pairwise transfers between tasks, which helps in identifying which tasks are likely to benefit from joint learning. In the second stage, it constructs a linear number of candidate sets of auxiliary tasks (in the number of all tasks) for each target task based on the initial estimations, significantly reducing the exponential number of potential auxiliary task sets. Thirdly, it employs a Multi-Armed Bandit (MAB) framework for each task, where the arms correspond to the performance of candidate auxiliary sets realized as multiple output neural networks over train-test data set splits. To enhance efficiency, BandiK integrates these individual task-specific MABs into a multi-bandit structure. The proposed multi-bandit solution exploits that the same neural network realizes multiple arms of different individual bandits corresponding to a given candidate set. This semi-overlapping arm property defines a novel multi-bandit cost/reward structure utilized in BandiK.

We validate our approach using a drug-target interaction benchmark. The results show that our methodology facilitates a computationally efficient task decomposition, making it a scalable solution for complex multi-task learning scenarios.
\end{abstract}
\end{frontmatter}


\section{Introduction}
\label{s:Intro}

The availability of foundation models in multiple domains, starting with vision and culminating in linguistic tasks redefined the scope of multitask learning as omni-task learning towards artificial general intelligence~\citep{subramanian2018learning,bubeck2023sparks}. Still, detailed quantitative evaluations of multitask learning are frequently grappling with the presence of negative transfer effects, i.e., the intricate, complex pattern of beneficial-detrimental effects of certain tasks on others~\citep{zhang2022survey}. It is by no surprise as the nature of transfer learning is multi-factorial and transfer effects can be attributed to (1) shared data, (2) shared latent representations, and (3) shared optimization; thus transfer effects are contextual and depend on the sample sizes, task similarities, sufficiency of hidden representations, and stages of the optimization~\citep{chen2018gradnorm,sener2018multi,yu2020gradient,liu2019loss,meng2021multi,yang2023adatask,xin2022current,zhou2020task,wu2020understanding,galanti2022improved,standley2020tasks,fifty2021efficiently}.

An alternative paradigm suggests selecting the beneficial auxiliary task subsets for each target task or eliminating the detrimental auxiliary tasks; however, evaluation of candidate auxiliary task subsets can be computationally demanding and statistically leads to loss of power due to multiple hypothesis testing.

\begin{figure}[H]
\vspace{-5pt}
\centering
\includegraphics[width=1\linewidth, trim={0 0 0 35pt}, clip]{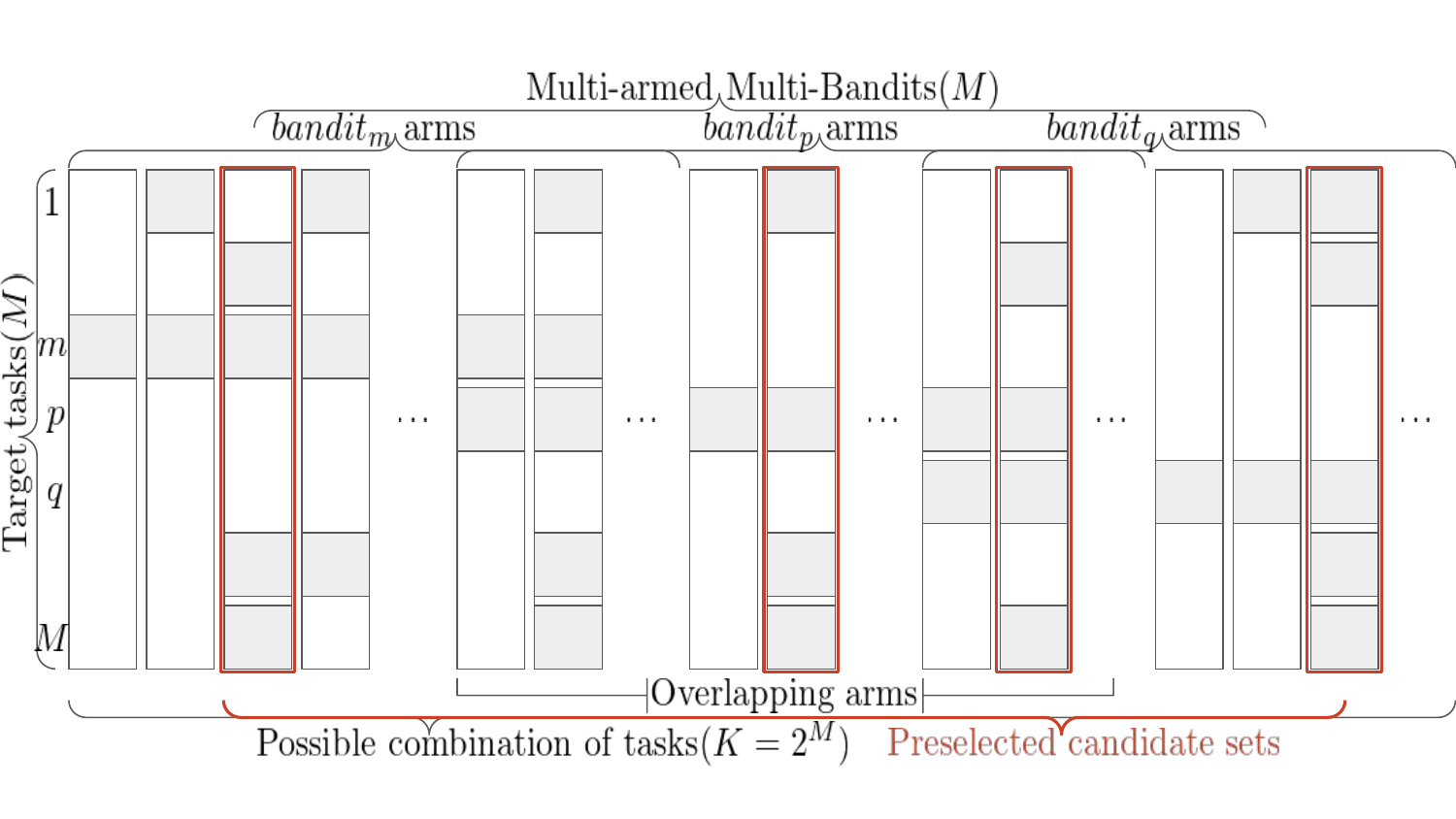}
\vspace{-15pt}
\caption{The multi-task subset selection problem as a multi-bandit: The selection of a multi-task neural network implies both the selection of a target set and their respective auxiliary task set.}
\label{fig:BandiK}
\end{figure}

To address these constraints, we introduce BandiK, a novel three-stage multi-task auxiliary task subset selection method using multi-bandits, where each arm pull evaluates candidate auxiliary sets by training and testing a multiple output neural network on a single random train-test dataset split (the problem setup is shown in Figure~\ref{fig:BandiK}). 

Firstly, BandiK estimates the pairwise transfers between tasks, which helps in identifying which tasks are likely to benefit from joint learning~\citep{xu2017demystifying,ben2010theory,pentina2017multi,zhou2020task}. In the second stage, it constructs a linear number of candidate sets of auxiliary tasks (in the number of all tasks) for each target task based on the initial estimations, significantly reducing the exponential number of potential auxiliary task sets~\citep{standley2020tasks}. Thirdly, it employs a Multi-Armed Bandit (MAB) framework including the tasks, where the arms correspond to the performance of candidate auxiliary sets realized as multiple output neural networks over train-test data set splits~\citep{gabillon2011multi,scarlett2019overlapping}. The major steps are summarized in Algorithm~\ref{alg:MBS4MTL}.

\begin{algorithm}
\caption{The main contribution of the BandiK multi-bandit method: pairwise transfer effect-based candidate (auxiliary) task sets as arms in bandits for each target task, and multi-task neural networks as semi-overlapping arms. }
\label{alg:MBS4MTL}
\begin{algorithmic}
\Procedure{\textbf{BandiK}}{}
    \State {\scshape Stage 1.}: Estimation of pairwise transfer learning effects between tasks and building positive and negative transfer graphs based on different tests for transfer
    \State {\scshape Stage 2.}: Construction of candidate auxiliary sets for each task by applying search methods in the graphs 
    \State {\scshape Stage 3.}: Definition and simulation of a multi-bandit using the adaptive GapE-V method; selection of best performing multi-task neural network for each target task.
\EndProcedure
\end{algorithmic}
\end{algorithm}

\textit{In this paper, we adopt the auxiliary task subset selection approach using hard parameter sharing networks and investigate (1) the construction methods of auxiliary sets, (2) the efficiency of a multi-bandit approach, and (3) the effects of the novel semi-overlapping arms in a multi-bandit.} We answer the following questions:
\begin{enumerate}[itemsep=0pt, parsep=0pt, topsep=0pt, partopsep=0pt]     \item[Q1] {\textit{Baseline performance}}: What are the best performances of single, pairwise, and multi-task learning scenarios per tasks? 
    \item[Q2] {\textit{Pairwise task landscape}}: What are the performances of pairwise target-auxiliary learning scenarios per tasks? 
    \item[Q3] {\textit{Uncertainty and heterogeneity of task performances}}: What are the variances of task performances? What is their heterogeneity and relation to task properties, such as task sample size?
    \item[Q4] {\textit{Heuristics for candidate auxiliary sets}}: What is the landscape of candidate auxiliary sets based on heuristics, such as greedy pairwise, filtered pairwise, pairwise-based transitive closures and cliques, incremental-decremental approaches?
    \item[Q5] {\textit{Multi-bandit dynamics}}: What are the distributions of pulls and convergence rates over individual bandits throughout learning, especially regarding their performance gaps and variances?
    \item[Q6] {\textit{Semi-overlapping arms}}: What are the effects of semi-overlapping arms between bandits, i.e., the massive multiple presence of shared networks and pulls?
    \item[Q7] {\textit{Performance of candidate auxiliary task sets}}: What are the performances of candidate auxiliary task sets, especially regarding their types and the availability of shared candidates (is there any cross-gain)?
    \item[Q8] {\textit{Nature of transfer}}: What are the implications of the results for the nature of transfer, i.e., transfer by shared samples, latent features or optimization?
    \item[Q9] {\textit{Applicability of foundation models}}: What are the implications for foundation models in special quantitative domains, e.g., in drug-target interaction prediction?
\end{enumerate}

We demonstrate our methods on a drug-target interaction (DTI) prediction problem~\citep{wang2022profiling}.

\section{Related works}

MABs are successfully applied in hyper-parameter optimization and neural architecture search, though we go a step deeper: a pull of an arm corresponds to a single train-test split evaluation in our case (see e.g.,~\citep{guo2019autosem}). The selection of multiple tasks for joint learning motivated a series of sequential learning methods~\citep{lugosi2009online}. Combinatorial multi-armed bandits (MABs) allow the pull of multiple arms simultaneously resulting in an aggregate reward of individual arms~\citep{cesa2012combinatorial}. Notably, the 'top-$k$' extension considers the subsets up to size $k$ and allows variants whether the rewards for the individual arms are available or only their non-linear aggregations~\citep{rejwan2020top,agarwal2022stochastic}. Another relevant extension to our formalization with multiple auxiliary task subset selection problem for each task is the multi-bandit approach, also allowing overlap between the arms~\citep{gabillon2011multi},~\citep{scarlett2019overlapping}. Finally, recent MAB extensions investigated the use of task representations and learning task relatedness~\citep{du2023multi,mukherjee2024multi,sessa2024multitask}. 

\section{Problem setup}
\label{sec:problem}

Consider a Multi-Task Learning problem with $M$ tasks, in which one should decide among $K=2^{M-1}$ options for auxiliary task sets for each $\task_m$ to reach optimal performance.
($2^{M-1}$ being the number of possible sets surely containing $\task_m$, which number we propose to significantly reduce in the first two stages of BandiK.)
This problem can be formulated as the best arm identification over $M$ multi-armed bandits with $K$ arms each.
In the setting each target task corresponds to a multi arm bandit and each set in the power set of tasks containing the target $\task_m$ is considered an arm.
(We use indices $m$, $p$, $q$ for the bandits and $k$, $i$, $j$ for the arms.)
Pulling an arm induces the training of a neural network on this set of tasks, and we are asked to recommend an arm for each bandit after a given number of pulls (budget). 

This setting is a special case of the multi-bandit best arm identification problem \cite{gabillon2011multi}. 
To acquire the reward of an $\arm_{mk}$ of $\bandit_m$, we train a neural network on the tasks covered by $\arm_{mk}$ (always containing at least $\task_m$ itself) that we denote $\armset_{mk}$. 
Notice that for a given task subset $\T$,
 at pulling any $\arm_{mk}$ with $\armset_{mk}=\cal{T}$ of $\bandit_m$ corresponding to target $\task_m\in\T$,
 an identical network can be constructed,
 just there are $|\T|$ variations which is the target task in $\T$ and which is the (complement) auxiliary task set.
Being the target task or an auxiliary one is only a matter of evaluation.
Thus, if $\armset_{mk}=\armset_{pi}$ 
 the cost of training the underlying network is incurred only once for $\arm_{mk}$ and $\arm_{pi}$,
 however, the corresponding rewards, depending on the target tasks ($\task_m$ vs. $\task_p$), are different.
Hence, we call such $\arm_{mk}$ and $\arm_{pi}$ \emph{semi-overlapping} arms
 (as opposed to (fully) overlapping arms with the same reward, e.g., in \cite{scarlett2019overlapping}).
Therefore, in our case joint optimization of the auxiliary task selection problem is not only reasonable for optimal resource allocation among the bandits,
 but more importantly each $\bandit_p$ having a semi-overlapping $\arm_{pi}$ with $\arm_{mk}$ can also make an update about the arm's reward whenever $\arm_{mk}$ is pulled in $\bandit_m$,
 because the neural network trained once can be evaluated for any included task,
 so the cost of the reward has to be paid only once.

Following the notation of \cite{gabillon2011multi}, let $M$ be the number of bandits and $K=2^{M-1}$ be the number of possible arms of each bandit. Each $\arm_{mk}$ of a $\bandit_m$ is characterized by a distribution $\nu_{mk}$ bounded in $[0, 1]$ with mean $\mu_{mk}$.
We denote by $\mu^*_m$ the mean and $k^*_m$ the index of the best arm of $\bandit_m$. 
In each $\bandit_m$, we define the gap for each arm as $\Delta_{mk} = \lvert \max_{j \neq k} \mu_{mj} - \mu_{mk} \rvert$.

At each round of the game $t = 1, \dots, n$, the forecaster pulls a bandit-arm pair $I(t)=(m,k)$ and observes a sample drawn from each distribution in 
$\{ \nu_{pi} : \armset_{pi} = \armset_{mk}\}$ 
independent of the past.
Let $T_{mk}(t)$ be the number of times that
 a sample from $\nu_{mk}$ has been observed
 by the end of round $t$, meaning $T_{mk}(t)=T_{pi}(t)$ throughout the whole game if $\armset_{mk} = \armset_{pi}$.
Let us note that, in contrast of the non-overlapping case,
 the final number $n$ of rounds is not equal to the sum of $T_{mk}(n)$'s for all $m$ and $k$, 
 leading to a massive relative budget increase.
We use 
 the adaptive GapE-V algorithm of \cite{gabillon2011multi}
 with the same values
 $\hmu_{mk}(t)$,
 $\hDe_{mk}(t)$,
 $\hsi_{mk}^2(t)$
 \footnote{Note that there is a missing $T_{mk}(t)/(T_{mk}(t)-1)$ factor in the definition of the unbiased sample variance $\hsi^2_{mk}(t)$ in \cite{gabillon2011multi}.}%
 , $B_{mk}(t)$,
 and $\hH^\sigma(t)$
 as derived in \cite{gabillon2011multi}.

In each $t$ time step, an $\arm_{mk}$ is pulled and an $f(t)=f(D_{\rm train}(t), \T(t))$ hard parameter sharing neural network is trained, from random initialization, to predict the $\T(t)=\armset_{mk}$ set of tasks, utilizing a random train split of the data.
Monte Carlo Subsampling \cite{shao1993linear} is used to split the dataset into $D_{\rm train}(t)$ and $D_{\rm test}(t)$ in each round, with sizes $80\%$ and $20\%$, respectively.
All $\bandit_p$ corresponding to a $\task_p\!\in\!\T(t)$ receives a sample $X_{pi}(t)$ by evaluating the performance of the network separately:
$X_{pi}(t)=L_{p}(f(t), D_{\rm test}(t))$, where $i$ is chosen so that $\armset_{pi} = \T(t)$.
Monte Carlo Cross Validation has been shown asymptotically consistent \cite{shao1993linear}.
Let $L_m(\T)=L_m(f(D_{\rm train}(u), \T), D_{\rm test}(u))$ denote the loss corresponding to $\task_m$ of a network trained and evaluated on a random data split, to predict the tasks in $\T$.
$L$ can be any standard loss function, but throughout this paper we use the Area Under the Receiver Operating Curve (AUROC, $L^{AUR}$)
and the Area Under the Precision-Recall Curve (AUPR, $L^{AUP}$) metrics.


\section{Data and methods}

\subsection{Data set}

Following~\cite{wang2022profiling}, we use the NURA-2021 data set binarized as 'strong binder' versus other labels. It includes $22$ targets and $31006$ compounds.
Since completely random train/test splits suffer from the compound series bias, leading to overoptimistic performance estimations~\cite{mayr18}, we utilize a more realistic, scaffold-based train/test split in the spirit of~\cite{simm21}, which resulted in $6441$ scaffolds.

\subsection{Models and computations}

Every time a network train is mentioned, the SparseChem ~\citep{arany2022sparsechem} model, a multiple output MLP is used.
Architecture and other hyperparameters were determined by grid search and apart from the number of neurons in the last layer these values were always unchanged.
Consequently, ADAM optimizer was used with a learning rate of $10^{-4}$ and weight decay of $10^{-6}$. 
Trainings run for $25$ epochs with $10\%$ batch ratio.
Following the $32000$ neuron-wide input layer, and two $1000$ neuron-wide hidden layers, the width of the output layer was always~$|\T|$.
ReLU activation and $0.7$ dropout ratio were used in the hidden layers. Sigmoid and $0.2$ dropout in the last layer. 

\subsection{BandiK}
\label{sec:bandig_algo}
We introduce BandiK, a novel three-stage  method using multi-bandits for solving the multi-task auxiliary task subset selection problem described in Section \ref{sec:problem}.

\subsubsection{Estimating pairwise transfer and building transfer graphs}
\label{sec:estimating_pairwise}
In the first stage, we discover multi-task transfer effects, by training on a number of base case candidate task sets.
There are four main training scenarios: single task learning (STL, $\T_m=\{m\}$), pairwise learning (PW, $\T_{pq}=\{p, q\}$), full multitask learning (FMTL, $\T_{F}=\{1, \dots, M\}$) and leave-one-out learning (LOO, $\T_{m-}=\T_{F} \setminus \{m\}$).
We train all possible scenarios on $500$ random dataset splits to acquire a sample of the network performances. 

Using the results of the base cases, we construct directed positive tranfer graphs ($P^{metric, test}$) and directed negative tranfer graphs ($N^{metric, test}$).
The existence of a $(p,q)$ directed edge in a given graph indicates that $\task_p$ has a (positive or negative) transfer effect on $\task_q$.
To construct the positive transfer effect matrices we check if PW options perform better than STL.
The edges of negative transfer matrices are derived from comparing LOO to FMTL, e.g., if a $task_q$ performs better with $\T_{p-}$ than trained with $\T_F$, it means that $\task_p$ has a clear individual negative transfer effect on $\task_q$.

\begin{enumerate}
    \item[-] $P^{AUP, diff}_{pq} = 1 \iff L^{AUP}_q(\T_{pq}) - L^{AUP}_q(\T_{q}) > 0$

    \item[-] $P^{AUR, diff}_{pq} = 1 \iff L^{AUR}_q(\T_{pq}) - L^{AUR}_q(\T_{q}) > 0$

    \item[-] $N^{AUP, diff}_{pq} = 1 \iff L^{AUP}_q(\T_{p-}) - L^{AUP}_q(\T_{F}) > 0$

    \item[-] $N^{AUR, diff}_{pq} = 1 \iff L^{AUR}_q(\T_{p-}) - L^{AUR}_q(\T_{F}) > 0$

\end{enumerate}

In addition to using the difference to indicate  better performance, we propose two kinds of significance tests to assess transfer effect: 
\begin{enumerate}
    \item[-] t-test: $P^{AUP, tt}$, $P^{AUR, tt}$, $N^{AUP, tt} N^{AUR, tt}$
    \item[-] nemenyi-thrd: $P^{AUP, nem}$, $P^{AUR, nem}$, $N^{AUP, nem}$, $N^{AUR, nem}$
\end{enumerate}
where the existence of a $(p, q)$ edges of the graphs in the t-test list are calculated using Student t-test with $p=0.05$, e.g., if the hypothesis $h_0: L^{AUP}_q(\T_{pq}) \leq 
L^{AUP}_q(\T_{q})$ can be rejected.
Nemenyi-thrd: for each $\task_m$, firts, Friedman Chi$^2$ test is run  to check if there are signifficant differences between the STL, PW, FMT and LOO parformances (44 distributions per task) and if so this is followed by Nemenyi test to search pair-wise significant performance differences between .

\subsubsection{Construction of candidate auxiliary task sets}
\label{sec:candidate_generation}

In the second stage, using transfer graphs constructed in Stage 1., we propose to construct a fixed number of candidate sets for each $task_m$, by running different heuristic based graph search algorithms
to find connected tasks with positive/negative transfer. These heuristics closely resemble forward and backward methods used in the feature subset selection problem~\citep{al2020approaches}.
In the case of positive transfer the initial candidate set is always $\T'=\T_m$ which is inflated with newly found nodes.
When negative transfer graphs are searched, results are removed from the initial $\T'=\T_F$.

Candidate sets are generated using the following methods.

The final candidate sets genereted by {\scshape GenerateArms}$(\{$
$P^{AUP, diff}$,$P^{AUR, diff}$, $N^{AUP, diff}$, $N^{AUR, diff}$, 
$P^{AUP, tt}$,  $P^{AUR, tt}$  , $N^{AUP, tt}$,   $N^{AUR, tt}$,
$P^{AUP, nem}$, $P^{AUR, nem}$ , $N^{AUP, nem}$,  $N^{AUR, nem}\}, M)$

$398$ unique candidate sets are generated in Stage 2.
using $4$ search methods that utilize different suppositions about the transitivity of pairwise transfer effects.

\begin{enumerate}[itemsep=0pt, parsep=0pt, topsep=0pt, partopsep=0pt] 
    \item[S1] {{\scshape DirectNeighbours}}: the most conservative, choosing only the inward pointing star graph of the direct neighbors of $task_m$
    \item[S2] {{\scshape Transitive}}: supposes that the pairwise transfer effects are fully transitive and so all the $task_p$ nodes in the graph with a connected (directed) path from $task_p$ to $task_m$ are chosen   
    \item[S3] {{\scshape FilteredTransitive}}: supposes that the dense transfer effect graph has a sparse generator, therefore, before running S2 it reduces the graph to its maximum weight spanning tree 
    \item[S4] {{\scshape Clique}}: supposes that the pairwise transfer is not only transitive, but should also be symmetric, so the biggest subset of fully connected nodes containing $task_m$ is chosen as the candidate set
\end{enumerate}

\subsubsection{Multi-Armed Bandit (MAB)} 
In the final stage, the problem of task-wise auxiliary task set selection is mapped to a Multi-Armed Multi-Bandit, where each task has a corresponding bandit and the arms give performance estimations of candidate auxiliary sets.
The $276$ base cases from Stage~1. and $398$ Stage~2. candidate sets give us a total of $674$ semi-overlapping arms.
In this stage the A-GapE-V algorithm \cite{gabillon2011multi} is run with a total of $22$ bandits and $398$ semi-overlapping arms, where the arm set of $\bandit_m$ is chosen to be the set of candidate sets containing $\task_m$.
After the termination of the algorithm each bandit/task is presented with the best arm/auxiliary task set.

\section{Results}

We present results about the multi-task properties of the DTI prediction problem, the properties of the multi-bandit architecture, its dynamics, and its performance.

\subsection{Domain characterization}

\begin{figure}[th]
\centering
\includegraphics[width=\linewidth]{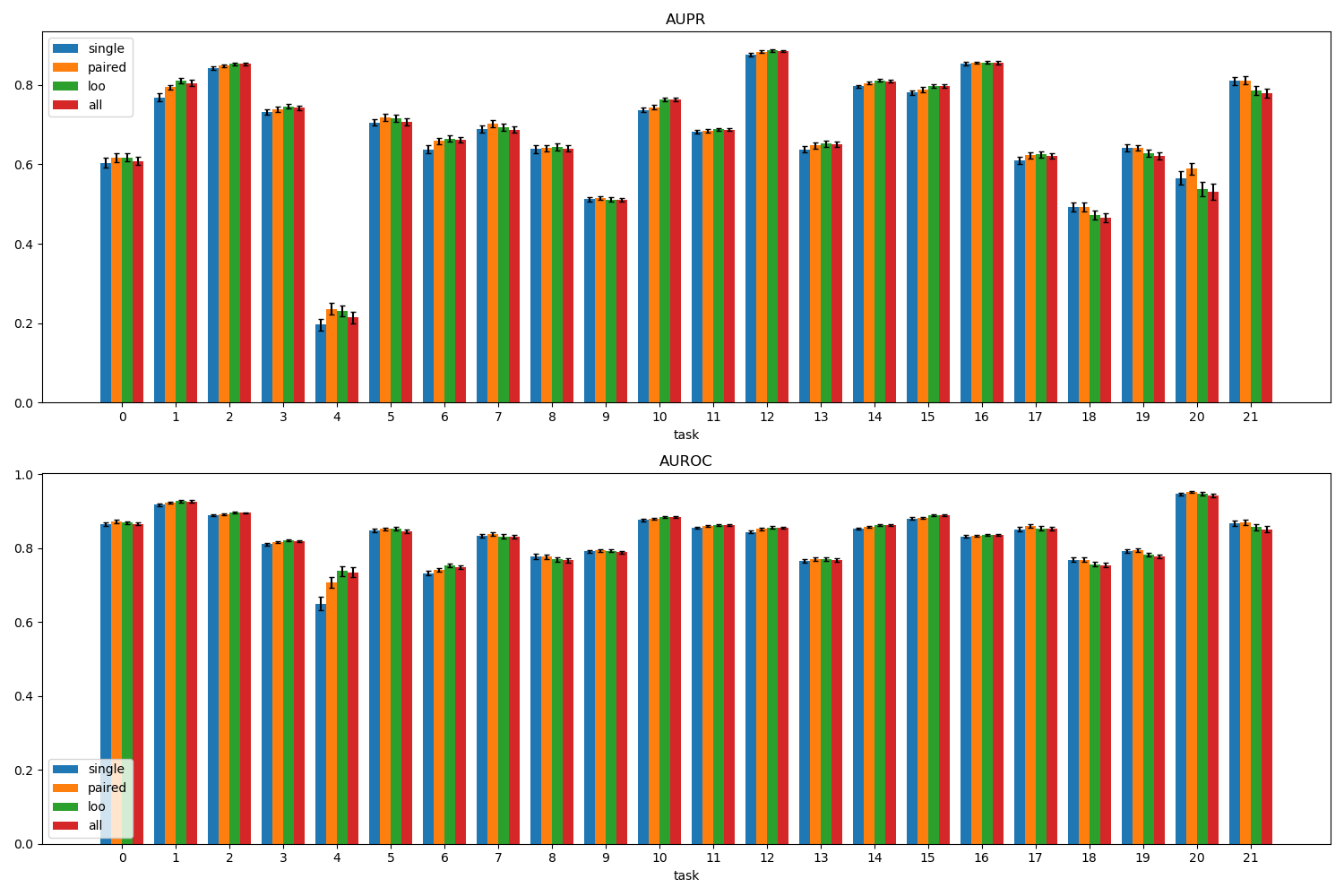} 
\caption{Task-by-task comparison of AUPR performances of the single-task (horizontal axis), multi-task and the best and worst of the pairwise and leave-one-out cases (vertical axis).\\}
\label{fig:DTI_Performance_landscape}
\end{figure}

We explored the performance landscape of the DTI domain and the possibility of multi-task improvements by estimating the base case scenarios of Section~\ref{sec:estimating_pairwise} (see Figure~\ref{fig:DTI_Performance_landscape}) .

\begin{figure}[hb]
\vskip 0pt
\centering
\includegraphics[width=\linewidth]{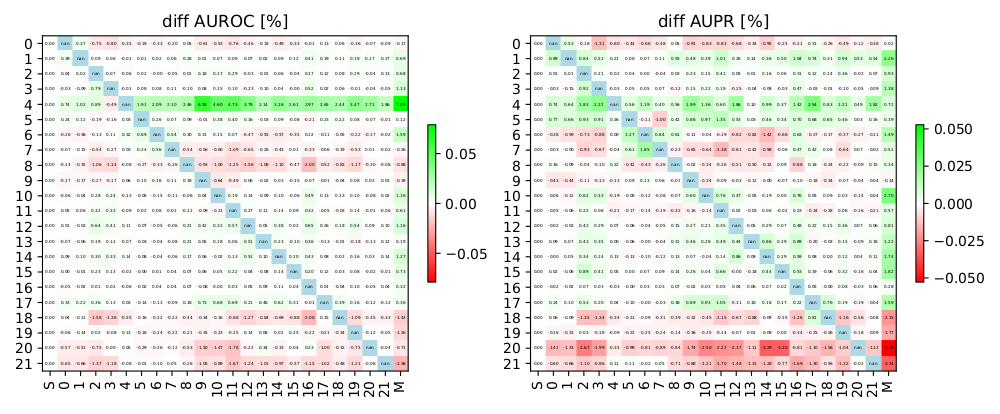} 
\caption{The pairwise transfer effects between tasks (difference of the AUROC and AUPR metrics, between the paired and single-task cases) The differences for the full multi-task case are also included in the column denoted by 'M'.
\\}
\label{fig:Pairwise_landscape}
\vskip 10pt
\end{figure}

The detailed pairwise transfer effects are illustrated in Figure~\ref{fig:Pairwise_landscape}.

\begin{figure}[ht]
\vskip -3pt
\centering
\includegraphics[width=0.7\linewidth]{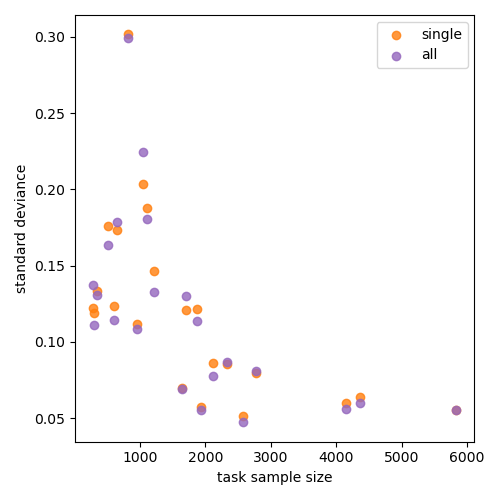}
\caption{The estimated standard deviations and the sample sizes for the tasks (estimates based on 500 data splits in Monte Carlo CV).\\}
\label{fig:Performance_variances_versus_task_samples}
\end{figure}

The variances of the performance metrics have a major impact on the number of necessary train-test data splits for reliable performance estimation and model selection. Figure~\ref{fig:Performance_variances_versus_task_samples} shows the sample sizes and the standard deviations of the tasks, which are a major factor determining their variances.

\subsection{Static properties of BandiK}

\begin{figure}[ht]
\vskip -3pt
\centering
\includegraphics[width=\linewidth]{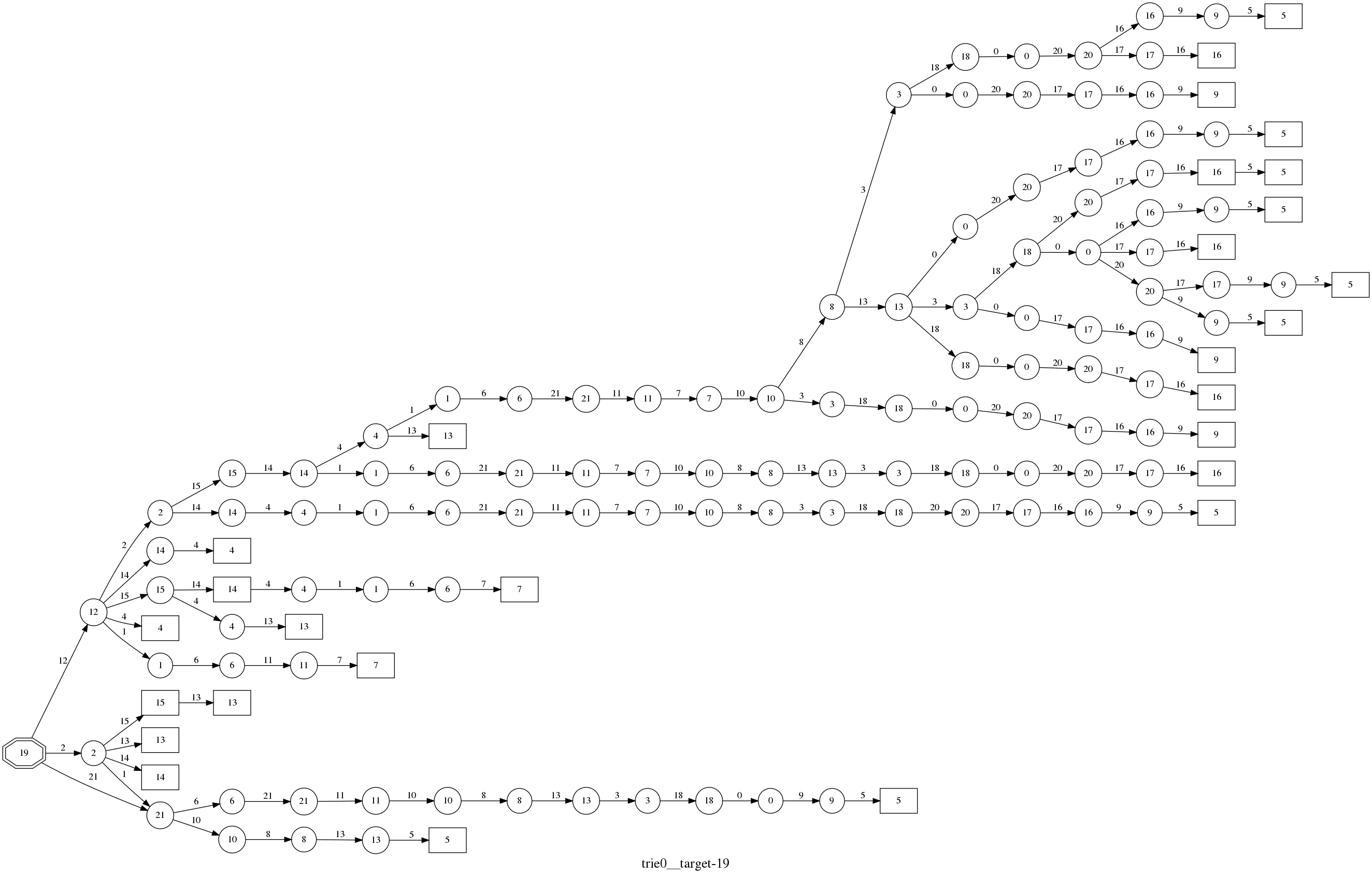} 
\caption{The candidate auxiliary task subsets for a given target task (task 19) illustrated in a trie. The octagon-shaped node represents the root task from which the graph searches were started, rectangular nodes denote leaf nodes (i.e. endings of candidate sets).
\\}
\label{fig:CandidatesInTrie}
\vskip 2pt
\end{figure}

\begin{figure}[b]
\vskip -1pt
\centering
\includegraphics[width=0.65\linewidth, angle=270]{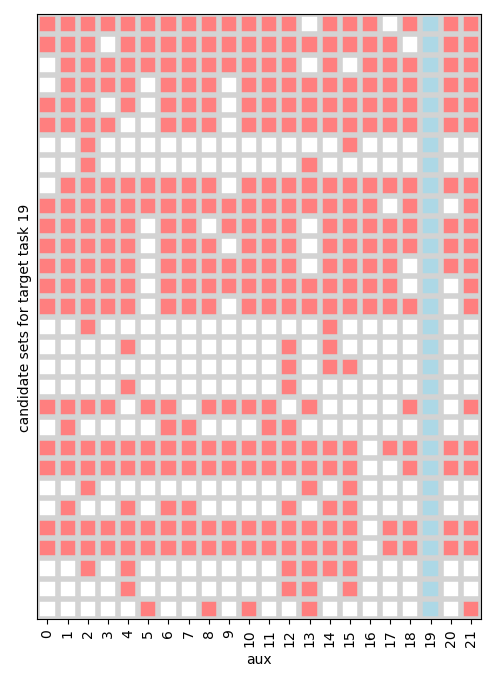}
\caption{Candidate auxiliary task subsets (i.e. neural networks) in a matrix for target 19 (columns correspond to sets/networks, rows correspond to auxiliary tasks.)\\
}
\label{fig:CandidatesInMatrix}
\vskip 0pt
\end{figure}

\begin{figure}[ht]
\centering
\includegraphics[width=0.9\linewidth]{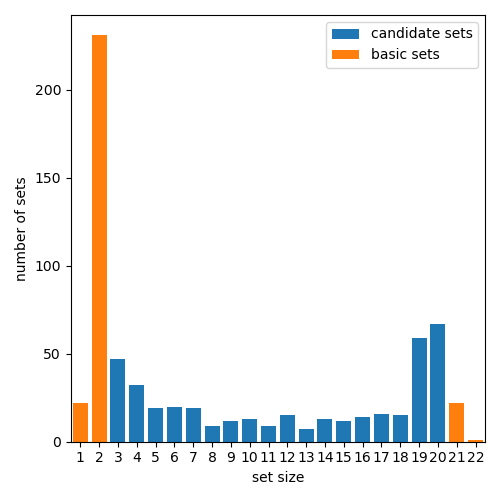}
\caption{Histogram of the size of the candidate auxiliary task subsets. Blue represents the heuristic candidate sets, while orange denotes the base-cases: single, pairwise, leave-one-out and full multi-task.
\\
}
\label{fig:HistogramOfCandidateSetsizes}
\end{figure}

The generated candidate auxiliary subsets (see Section~\ref{sec:candidate_generation}) for a given task are illustrated in Figure~\ref{fig:CandidatesInTrie} and~\ref{fig:CandidatesInMatrix}. A histogram of the sizes of all  generated candidate auxiliary subsets are shown in Figure~\ref{fig:HistogramOfCandidateSetsizes}.

\subsection{Dynamics of BandiK}

\begin{figure}[ht]
\centering
\includegraphics[width=\linewidth]{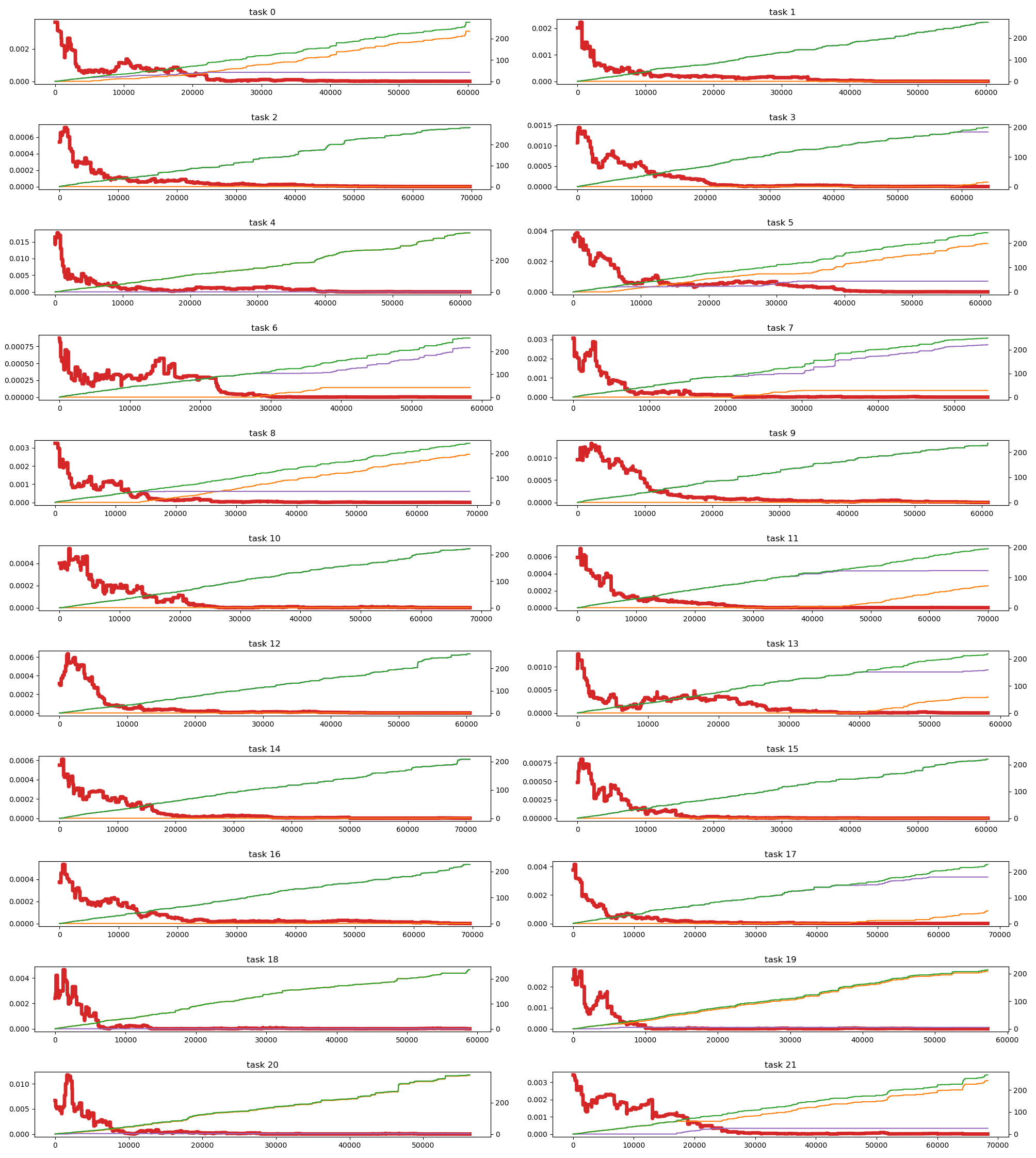}
\caption{Performances (simple regret, red line) of individual bandits corresponding to the target tasks throughout learning and the number of own (orange) and induced (purple) pulls on the best arm and their sum (green).
\\}
\label{fig:LearningDynamicsOfBanditsInBandiK}
\end{figure}

The performances of the individual bandits corresponding to the target tasks are shown in Figure~\ref{fig:LearningDynamicsOfBanditsInBandiK}. The final best arms are used as references and the simple regret, the selection error of a non-optimal arm in a given bandit, and the selection error of any non-optimal arm in the multi-bandit~\citep{gabillon2011multi,scarlett2019overlapping} are shown.

\begin{figure}[ht]
\centering
\includegraphics[width=0.9\linewidth]{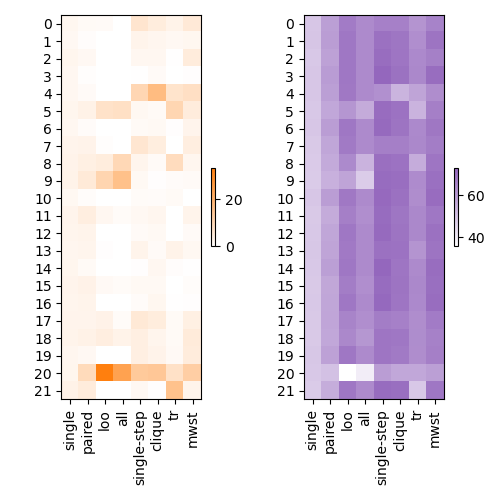}
\caption{Number of pulls by the type of candidate sets for each bandit (corresponding to tasks); numbers are normalized by the number of candidate sets of each type. Left: pulls directly initiated by the given arm (own pulls), left: pulls from which the arm received a sample (induced pulls).
\\}
\label{fig:PullsWithOverlapPulls}
\end{figure}

We also investigated the effects of the semi-overlapping arms in BandiK, i.e., the effect of shared neural networks between the bandits in the multi-bandit structure. Figure~\ref{fig:PullsWithOverlapPulls} shows the number of pulls per bandits and the ratio of own pulls versus other pulls originating from such semi-overlapping arms.

\begin{figure}[b]
\centering
\includegraphics[width=\linewidth]{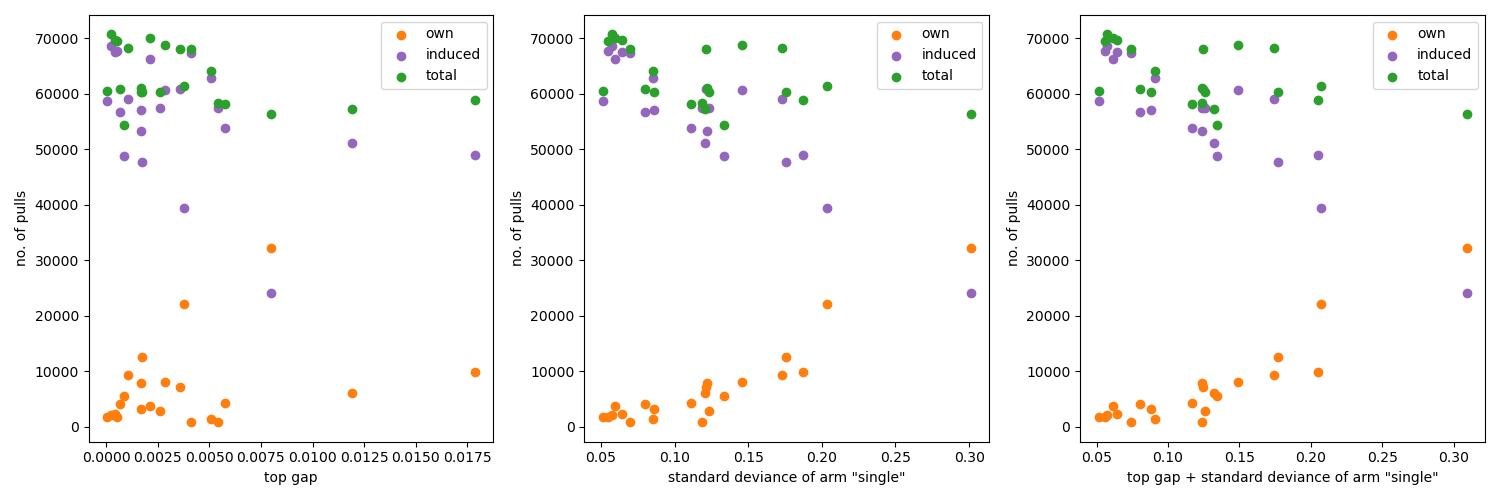}
\caption{The relation between the number of pulls, gaps, and variances (the pulls are separated according to their origin corresponding to its own arm or to a semi-overlapping arm).\\}
\label{fig:PullsVersusGapsAndVariances}
\end{figure}

The learning process in the GapE-V method is driven by the gaps between the expected values of the arms, specifically by the gap between the two top arms, and their variances. Figure~\ref{fig:PullsVersusGapsAndVariances} illustrates the relation between the number of pulls, gaps, and variances. The pulls are separated according to their origin: coming from the pull of the arm of the actual bandit or from the pull of a semi-overlapping arm of another bandit. Table~\ref{t:OwnVersusTotalPullsPerTasks} shows the ratios of own and total pulls for each bandits/tasks.

\begin{table}[!ht]
    \centering
    \begin{tabular}{cccc}
    \hline
        task & own pulls & induced pulls & ratio of own pulls \\ \hline
0 & 12630 & 60273 & 20.95\% \\
1 & 2848 & 60293 & 4.72\% \\
2 & 2263 & 69719 & 3.25\% \\
3 & 1344 & 64094 & 2.10\% \\
4 & 22035 & 61406 & 35.88\% \\
5 & 7805 & 61119 & 12.77\% \\
6 & 808 & 58277 & 1.39\% \\
7 & 5533 & 54378 & 10.18\% \\
8 & 8136 & 68726 & 11.84\% \\
9 & 4134 & 60873 & 6.79\% \\
10 & 872 & 68127 & 1.28\% \\
11 & 3716 & 69973 & 5.31\% \\
12 & 1721 & 60494 & 2.84\% \\
13 & 4379 & 58220 & 7.52\% \\
14 & 2076 & 70694 & 2.94\% \\
15 & 3215 & 60287 & 5.33\% \\
16 & 1804 & 69516 & 2.60\% \\
17 & 7122 & 68042 & 10.47\% \\
18 & 9867 & 58904 & 16.75\% \\
19 & 6176 & 57290 & 10.78\% \\
20 & 32224 & 56309 & 57.23\% \\
21 & 9292 & 68301 & 13.60\% \\
\hline
TOTAL & 150000 & 1385315 & 10.83\% \\
    \end{tabular}
    \caption{The number of own and total pulls and their ratio for each task.\\}
    \label{t:OwnVersusTotalPullsPerTasks}
\vskip -5pt
\end{table}

\subsection{Output of BandiK}

\begin{figure}[ht]
\centering
\includegraphics[width=\linewidth]{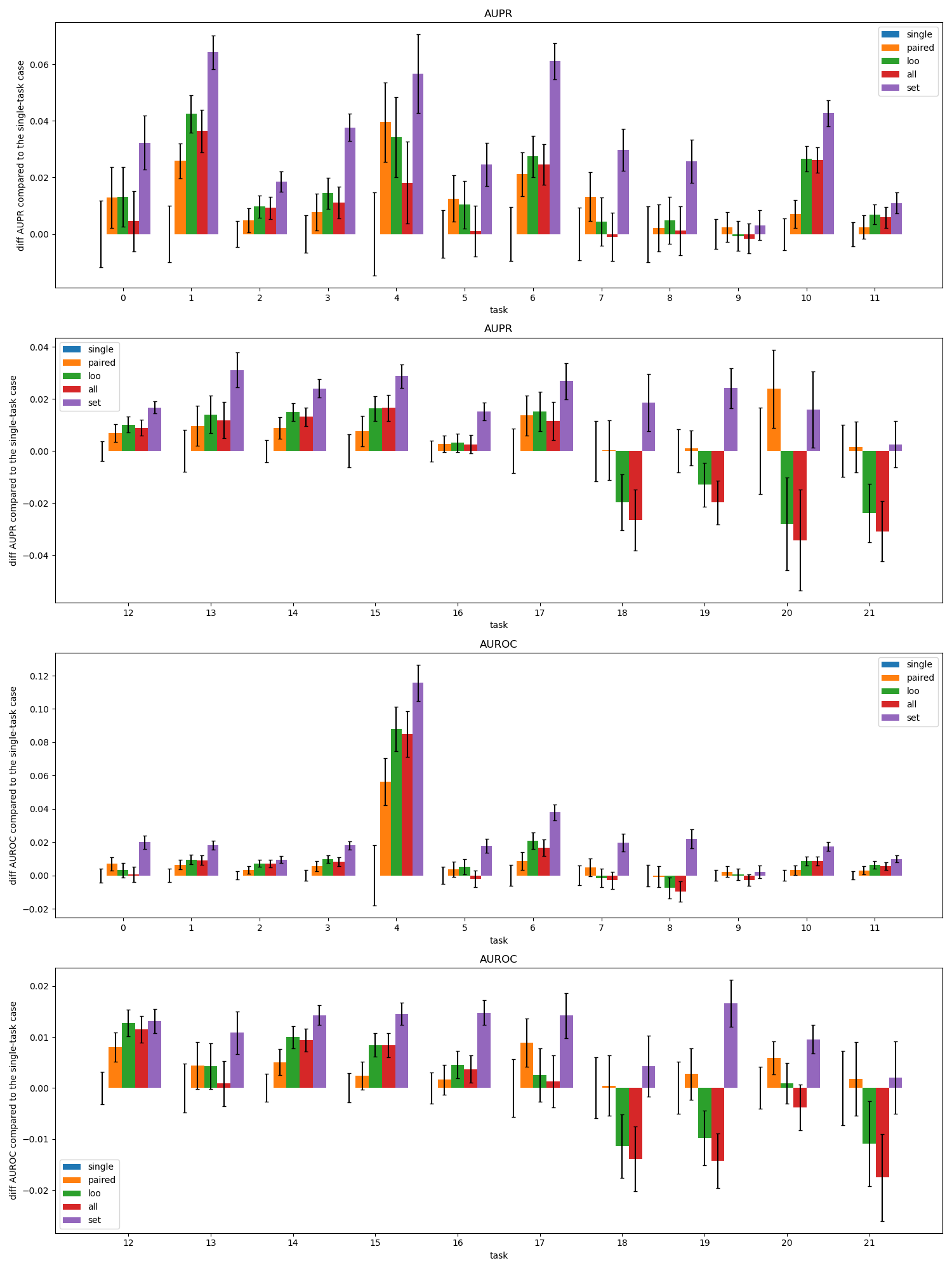}
\caption{Best performances of various types of AUPR performances of candidate auxiliary task subsets for each task compared to the single-task case.\\}
\label{fig:PerformancesAUPR_AUROC}
\end{figure}

The best performances of various types of candidate auxiliary task subsets for each task are shown in Figure~\ref{fig:PerformancesAUPR_AUROC} for the AUPR and AUROC measures.

\begin{figure}[ht]
\centering
\includegraphics[width=\linewidth]{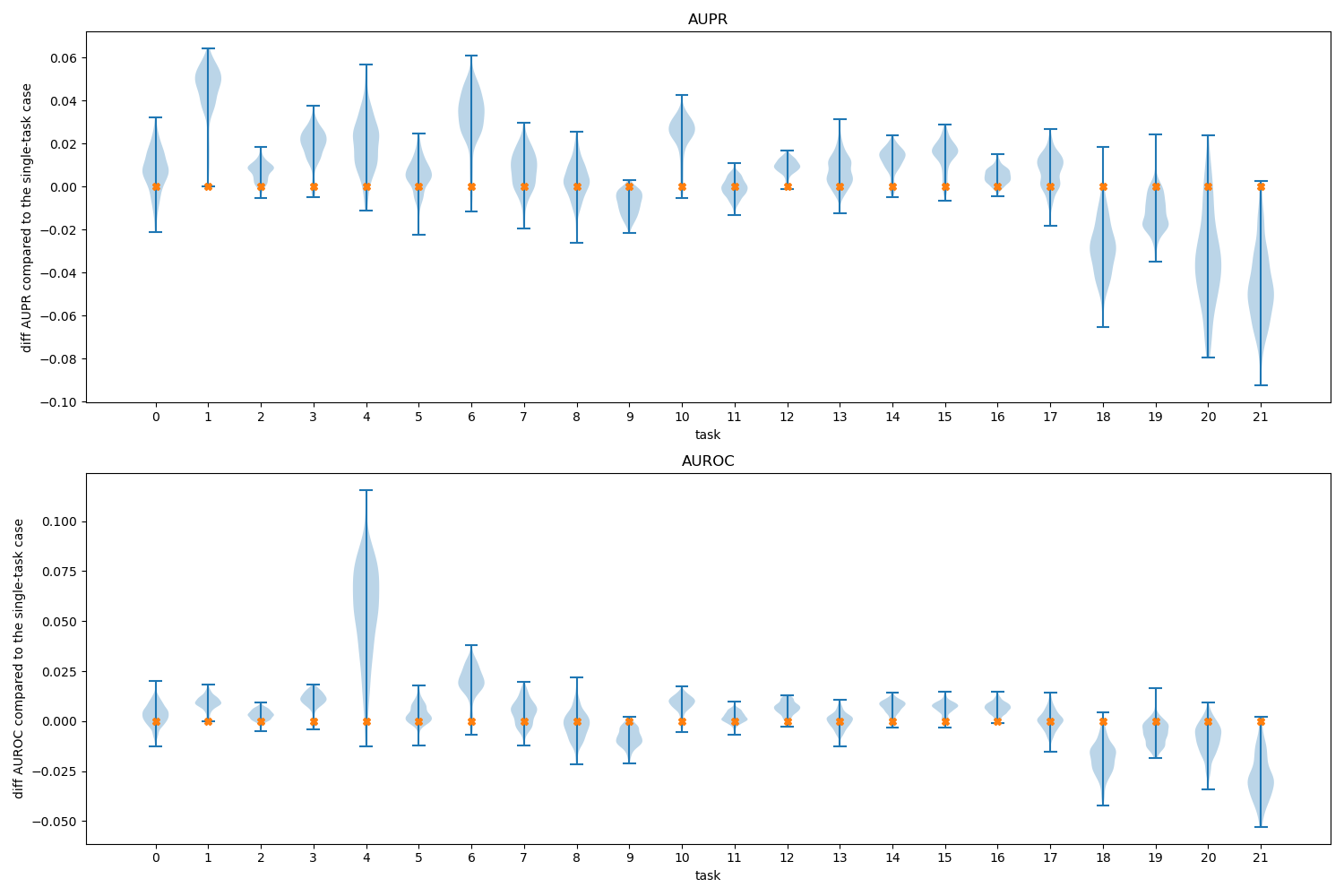}
\caption{Distribution of performances of arms (corresponding to cadidate auxiliary sets) for each multi-armed bandit for the AUPR metric.\\}
\label{fig:Violin_AUPR}
\end{figure}

Figure~\ref{fig:Violin_AUPR} shows the distributions of the performances of the arms for each multi-armed bandit (corresponding to tasks), again using the single-task case as reference point.

\begin{figure}[ht]
\centering
\includegraphics[width=\linewidth]{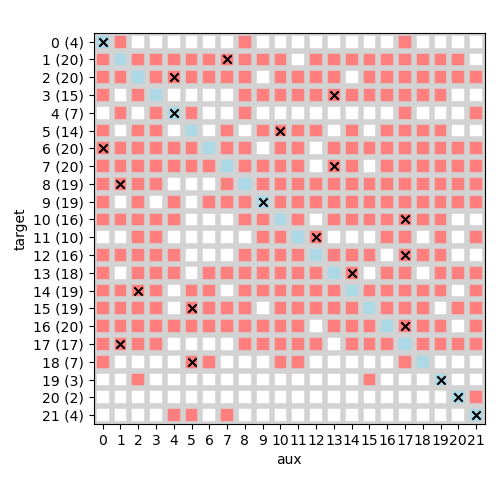}
\caption{The best candidate auxiliary subset for each task. Rows correspond to target tasks, cross sign indicates the origin (the generator) of the candidate auxiliary subset, red denotes the members of the set (network). Set sizes are displayed in parentheses.
\\}
\label{fig:BestNetsWithOrigin}
\vskip 3pt
\end{figure}

The best candidate auxiliary subsets, are shown in Figure~\ref{fig:BestNetsWithOrigin}. Note that the cross signs in the diagonal indicate when the best network comes from its own candidate set. 

\begin{figure}[ht]
\centering
\includegraphics[width=\linewidth]{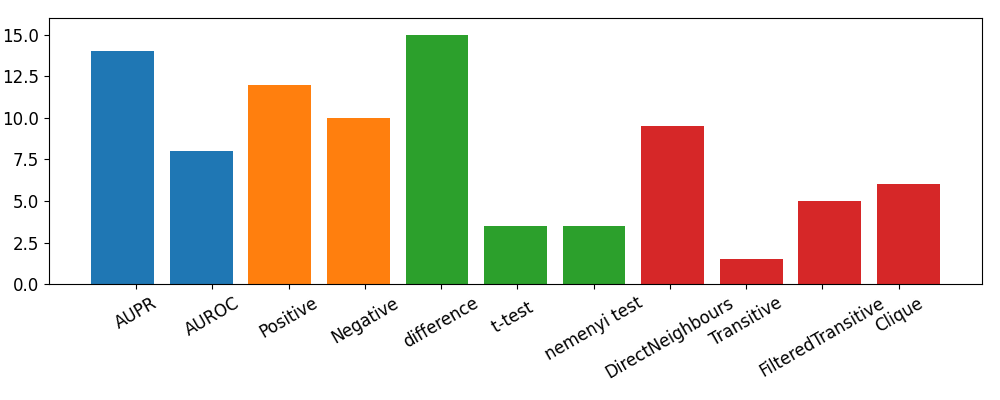}
\caption{The distribution of the best candidate auxiliary sets according to the origin of its generation. 
The first group correspond to the metric used for evaluation, the second group indicates the type of transfer  matrix used, and the third grouping is based on the test used for including edges in the graph (see \ref{sec:estimating_pairwise}).
The last group shows the chosen candidates as a proportion of search algorithm used (see \ref{sec:candidate_generation}).\\}
\label{fig:SourceOfBestsolution}
\vskip 1pt
\end{figure}

The distribution of the best candidate auxiliary sets according to the origin of candidate set generation is shown in Figure~\ref{fig:SourceOfBestsolution}.
The properties of the best candidate auxiliary sets are summarized in Table~\ref{table:chosen_lambdas}.

\begin{table}[t]
\vskip 1pt
  \caption{ Final selected candidate sets per task. The columns correspond to the metric used for evaluation, the type of transfer  matrix used, the test used for including edges in the graph (see \ref{sec:estimating_pairwise}) the search algorithm to choose candidates and finally the original task for which the candidate has been designed. If the first cell is empty it means that the same candidate set as the previous was found twice.}
  \label{table:chosen_lambdas}
  \centering
  \begin{tabular}{|l|rrrrr|}
    \hline %
    Task & Metric & Graph & Test & Search                        & Start Search  \\
    \hline %
     0 & AUPR  & + & diff & {\scshape Neighbours}   & 0  \\
     1 & AUROC & - & diff & {\scshape Clique}             & 7  \\
     2 & AUPR  & - & tt   & {\scshape Neighbours}   & 4  \\
     3 & AUPR  & + & nem  & {\scshape Transitive}         & 13  \\
     4 & AUROC & - & diff & {\scshape Filtered} & 4  \\
     5 & AUPR  & - & diff & {\scshape Neighbours}   & 10  \\
     6 & AUPR  & - & diff & {\scshape Clique}             & 0  \\
       & AUPR  & - & diff & {\scshape Clique}             & 3  \\
     7 & AUROC & - & diff & {\scshape Clique}             & 13  \\
     8 & AUROC & + & tt   & {\scshape Filtered} & 1  \\
     9 & AUROC & - & diff & {\scshape Clique}             & 9  \\
    10 & AUPR  & + & diff & {\scshape Neighbours}   & 17  \\
    11 & AUROC & + & nem  & {\scshape Filtered} & 12  \\
    12 & AUROC & + & diff & {\scshape Neighbours}   & 17  \\
    13 & AUPR  & - & diff & {\scshape Neighbours}   & 14  \\
    14 & AUPR  & + & diff & {\scshape Neighbours}   & 2  \\
    15 & AUPR  & - & diff & {\scshape Clique}             & 5  \\
    16 & AUPR  & - & tt   & {\scshape Filtered} & 17  \\
    17 & AUPR  & + & tt   & {\scshape Filtered} & 1  \\
       & AUPR  & + & nem  & {\scshape Transitive}         & 1  \\
    18 & AUROC & + & nem  & {\scshape Filtered} & 5  \\
       & AUROC & + & nem  & {\scshape Neighbours}   & 5  \\
    19 & AUPR  & + & diff & {\scshape Clique}             & 19  \\
    20 & AUPR  & + & diff & {\scshape Neighbours}   & 7  \\
    21 & AUPR  & + & diff & {\scshape Neighbours}   & 21  \\
    \hline %
  \end{tabular}
\end{table}

\section{Discussion}

The exploration of the baseline performances (Q1), the single, pairwise, and multi-task learning scenarios indicated the possibility of improvements (Q1), see Figure~\ref{fig:DTI_Performance_landscape}. 

The pairwise task landscape (Q2) confirmed other studies about the existence of central auxiliary tasks, which are useful auxiliary tasks for many other tasks, tasks, which can utilize many other tasks, and the existence of tasks, which seems unrelated to all other tasks using a pairwise task-task similarity approach (see task $4$, $20$ Figure~\ref{fig:Pairwise_landscape}.
 
The variances of the task performances were surprisingly high, which could be related to the highly incomplete and missing-not-at-random property of the DTI domain; furthermore, the variances were highly varying (Q3), see Figure~\ref{fig:Performance_variances_versus_task_samples}. 

We explored a wide range of heuristics to generate a fixed list of candidate auxiliary subsets (Q4); however, their redundancy indicated a high-level consistency in the domain, see Figure~\ref{fig:CandidatesInTrie} and~\ref{fig:CandidatesInMatrix}.

The distributions of pulls and convergence rates over individual bandits throughout learning were in line with the theoretical expectations and confirmed the sufficiency of the applied simulation length (Q5), see e.g. Figure~\ref{fig:LearningDynamicsOfBanditsInBandiK} and~\ref{fig:PullsVersusGapsAndVariances}.

The novel feature of BandiK, the semi-overlapping arms between bandits, proved to be extremely useful, also in the training process (Q6). Figure~\ref{fig:PullsWithOverlapPulls} and Table~\ref{t:OwnVersusTotalPullsPerTasks} show the ratio of such overlap pulls (dominated by external pulls). 

The performances of candidate auxiliary task sets and the final results also confirmed the fundamental importance of the semi-overlapping nature of BandiK: Figure~\ref{fig:SourceOfBestsolution} shows that 18 of the 22 tasks have an externally generated best network (Q7).

The types of the best networks indicate that complex generation methods, including aspects of multi-task consistency, positive and negative tranfer, and filtering are all useful (Q8), see Table~\ref{table:chosen_lambdas} and Figure~\ref{fig:SourceOfBestsolution}. This suggests that shared latent features, and not shared data, are behind large parts of the transfer effects (we minimized the cross-optimization effects using sufficiently lengthy training). However, our results also indicate that in special quantitative domains, such as in drug-target interaction prediction, explicit modeling of latent features, specifically the modeling of drug-target mechanisms, could be essential to understand, detect, and avoid negative transfer (Q9)~\citep{huang2021moltrans,wang2022yuel,bai2023interpretable}.

The performances of candidate auxiliary task sets resulted in a significant improvement for each task, see Figure~\ref{fig:PerformancesAUPR_AUROC}.

\section{Conclusion}

We introduced BandiK, a novel three-stage multi-task auxiliary task subset selection method using multi-bandits, where each arm pull evaluates candidate auxiliary sets by training and testing a multiple output neural network on a single random train-test dataset split. To enhance efficiency, BandiK integrates these individual task-specific MABs into a multi-bandit structure, which coordinates the selection process across tasks and ensures a globally efficient task subset selection for all the tasks. The proposed multi-bandit solution exploits that the same neural network realizes multiple arms of different individual bandits corresponding to a given candidate set. This semi-overlapping arm property defines a novel multi-bandit cost/reward structure utilized in BandiK.

We validated our approach using a drug-target interaction benchmark. The results show that our methodology facilitates a computationally efficient task decomposition, making it a scalable solution for complex multi-task learning scenarios. These could include more advanced methods modeling the task dependency-independency structure, usage of task similarity metrics, which can cope with the high-level incompleteness, and the use of adaptive construction methods already applied in neural architecture search. The computational efficiency of BandiK could be a key to scale up the current approach using fixed candidate auxiliary task sets to a full-fledged Monte Carlo Tree Search method.



\begin{ack}
This study was supported by the European Union project RRF-2.3.1-21-2022-00004 within the framework of the Artificial Intelligence National Laboratory.
\end{ack}



\bibliography{MTLTransfer24oct}

\end{document}